\documentclass[manuscript,screen]{acmart}

\AtBeginDocument{%
  \providecommand\BibTeX{{%
    \normalfont B\kern-0.5em{\scshape i\kern-0.25em b}\kern-0.8em\TeX}}}

\setcopyright{none}
\copyrightyear{2023}
\acmYear{2023}
\acmDOI{XXXXXXX.XXXXXXX}

\renewcommand\footnotetextcopyrightpermission[1]{} 

\acmConference[Many Worlds of AI '23]{Intercultural Approaches to the Ethics of Artificial Intelligence}{April 26--28, 2023}{Cambridge, UK}

\begin{document}

\title{Towards a Praxis for Intercultural Ethics in Explainable AI}

\author{Chinasa T. Okolo}
\email{chinasa@cs.cornell.edu}
\orcid{0000-0002-6474-3378}
\affiliation{%
  \department{Computer Science}
  \institution{Cornell University}
  \streetaddress{350 Gates Hall}
  \city{Ithaca}
  \state{New York}
  \country{United States}}

\renewcommand{\shortauthors}{Okolo}

\begin{abstract}
Explainable AI (XAI) is often promoted with the idea of helping users understand how machine learning models function and produce predictions. Still, most of these benefits are reserved for those with specialized domain knowledge, such as machine learning developers. Recent research has argued that making AI explainable can be a viable way of making AI more useful in real-world contexts, especially within low-resource domains in the Global South. While AI has transcended borders, a limited amount of work focuses on democratizing the concept of explainable AI to the ``majority world”, leaving much room to explore and develop new approaches within this space that cater to the distinct needs of users within culturally and socially-diverse regions. This article introduces the concept of an intercultural ethics approach to AI explainability. It examines how cultural nuances impact the adoption and use of technology, the factors that impede how technical concepts such as AI are explained, and how integrating an intercultural ethics approach in the development of XAI can improve user understanding and facilitate efficient usage of these methods.
\end{abstract}

\ccsdesc[500]{Human-centered computing~Collaborative and social computing}
\ccsdesc[500]{Social and professional topics~Cultural characteristics}
\ccsdesc[500]{Computing methodologies~Machine learning}
\ccsdesc[500]{Computing methodologies~Artificial intelligence}

\keywords{Artificial Intelligence, Intercultural Ethics, Intercultural XAI, Explainability, HCI, XAI4D, Global South}


\maketitle

\section{Introduction}
\label{intro}

Recent research has argued that making AI explainable can make AI more useful in real-world contexts, especially within low-resource domains in the Global South \cite{okolo2022making}. Other research has also noted the concentration of AI research and development within the Western world \cite{chan2021limits, okolo2023inequity} and how AI systems primarily embed the cultural values and practices of people within these respective regions, alienating certain groups of users and causing cultural harm \cite{prabhakaran2022cultural}. While the development, use, and implementation of AI has transcended borders, a limited amount of work focuses on democratizing the concept of explainable AI to the “majority world” \cite{alam2008majority}, leaving much room to explore and develop new approaches within this space that cater to the distinct needs of users within this region. 

This article introduces the concept of an intercultural ethics approach to AI explainability. It examines how cultural nuances impact the adoption and use of technology, the factors that impede how technical concepts such as AI are explained, and how integrating an intercultural ethics approach in the development of XAI can improve user understanding and facilitate efficient usage of these methods. We first discuss what it means to explain, reviewing relevant literature within explainable AI and introducing intercultural ethics. We then highlight barriers that impact explainable AI. Next, we introduce the concept of using intercultural ethics to inform AI explainability, outlining steps for researchers interested in leveraging this approach. We conclude the paper by reflecting on the prospect of an intercultural ethics approach to XAI, the limitations of such an approach, and potential areas to build upon this work.

\section{Background}
\label{background}

\subsection{What does it mean to explain?}
The Cambridge dictionary defines the verb explain as a way "to make something clear or easy to understand by describing or giving information about it" \cite{camdicExplain}. Overall, explanations can make things clear and understandable by supplying context, information, and reasoning. Other ways to explain involve simplifying abstract concepts into more interpretable forms. When explaining, additional tools like analogies, diagrams, and examples can be used to convey information simply. To explain effectively, factors such as clear communication skills, understanding the explanation needs of the target audience, and an ability to adjust communication skills to adapt to target audiences' needs are key. Adequate explanation allows knowledge to transfer clearly and comprehensively. The concept of explaining is present in various fields and has most recently begun to gain traction in the application of technical and human-centered methods in machine learning (ML) and artificial intelligence (AI).

\subsubsection{Explainable AI}
As a result of the growing need for developers and end users to manage AI systems and understand the decisions produced by these systems, researchers established the field of explainable AI (XAI). XAI aims to improve AI systems' transparency, interoperability, and trustworthiness while enhancing how they are deployed ethically and held accountable for the impact they may cause. XAI consists of methods that enable humans to understand the predictions made by machine learning models \cite{arrieta2020explainable, hagras2018toward} and can be present in the form of algorithms, toolkits, and libraries or non-technical forms such as frameworks and design guidelines. Early research in this field can be traced back to the 1980s in work by Buchanan and Shortliffe \cite{buchanan1984rule} providing explanations of infection diagnoses to doctors from a rule-based system and work by Mitchell et al.~\cite{mitchell1986explanation} using explanations to provide domain experts with ways to generalize ML systems. XAI methods work across a range of data types and use cases, being leveraged for models trained on images and videos in computer vision \cite{jung2021towards, lee2021lfi, oh2021evet, patro2019u, selvaraju2017grad} to text and tabular data in natural language processing (NLP) \cite{ribeiro2018anchors, li2016visualizing, harbecke2018learning, jiang2019explore}. XAI methods can be classified as either pre-hoc explainability, where explanations are produced in tandem with model predictions, and post-hoc explainability, where explanations are produced separately from model predictions, with no effect on the predictive accuracy of the model as a whole \cite{xu2022learning, murdoch2019interpretable}.

A primary goal of XAI is to provide human-interpretable reasoning for the predictions and decisions produced by AI systems. Explanations can impact how people use and trust AI systems. Still, issues of bias pervade these systems \cite{benjamin2019race, noble2018algorithms, eubanks2018automating}, and many existing XAI methods are only interpretable to a small subset of technically-savvy users. With this in mind, there is a need to broadly study existing use cases of XAI and develop novel methods that ensure end users from various backgrounds (educational, regional, cultural, etc.) can comprehensively understand AI decisions and use explanations to leverage AI in their respective work effectively. Over the past decade, research has emerged that uses human-centered methods to critically examine XAI \cite{okolo2022making, ehsan2022human, alikhademi2021can}, evaluate XAI with human users \cite{liao2020questioning, jesus2021can, cheng2019explaining, kim2022hive, nguyen2022visual}, and develop novel XAI methods \cite{ehsan2019automated, broekens2010you, shen2020designing}. This growing body of work in human-centered XAI shows much promise in improving current XAI approaches and creating new ones. Improving the interpretability of XAI can possibly enhance user confidence, reduce bias, improve the identification of prediction errors, and allow users to retain their autonomy when deciding whether or not to act on AI predictions. Additionally, leveraging methods from the humanities can help make progress toward more interpretable XAI, improving its utility for a diverse set of users in global contexts. 

\subsection{Intercultural ethics}
Researchers should incorporate an intercultural ethics approach into AI development to understand the challenges posed by the widespread use of AI and shape novel approaches to AI explainability that account for a globally diverse user base. Evanoff introduces the concept of intercultural ethics and defines it as ``the process by which people from different cultures negotiate the norms that will govern relations between them at a variety of levels, including the interpersonal, intergroup and international" \cite{evanoff19966intercultural, evanoff2006integration, evanoff2020introducing}. More generally, intercultural ethics help guide human behavior when interacting with individuals or groups from cultures different from their own. In these respective interactions, cultural beliefs, practices, and perspectives are understood and respected, and ethical standards that are sensitive to cultural differences are applied. 


In the case of explainable AI, applying intercultural ethics can improve the efficacy of explanations and make AI systems, in general, more useful for end users with low levels of AI knowledge. This paper introduces and examines the concept of an intercultural ethics approach to AI explainability to unpack the existing challenges of XAI and to promote human-centered approaches to model interpretability. This work aims to foster new ground for explainability approaches that cater to a wide range of users and improve how these techniques convey the complexity behind model decision-making.    
\section{Barriers to Explanations}
\label{barriers}

In the case of technology development, use, and adoption, cultural nuances can have a substantial impact. Cultural nuances are demonstrated in how people in different cultures express their beliefs, customs, values, and behaviors. They affect many aspects of our lives, including how we communicate and interact with others in verbal and non-verbal ways, such as language, social cues, and gestures. If cultural nuances are understood, cross-cultural communication is more effective, improving the respect of cultural perspectives and practices, and preventing misunderstandings and conflict. Cultural nuances also affect the design and development of technology due to the expectations, preferences, and values of groups of people. These factors can also influence how people adopt and use technology due to social norms around trusting and actively relying on technical systems. Research within HCI and ICTD has demonstrated that there exist cultural differences regarding privacy \cite{naveed2022ask}, technology use \cite{kyriakoullis2016culture}, interface design \cite{alostath2009identifying}, and algorithmic fairness \cite{sambasivan2021re}, which could impact how ethical values are embedded into technical products. For example, in cultures where collective interests are prioritized over individual privacy and autonomy, device sharing amongst multiple users is common \cite{sambasivan2010intermediated, ahmed2013ecologies, burrell2010evaluating}, but technical interfaces that account for this type of usage in such contexts are not the norm despite the many challenges noted by researchers \cite{sambasivan2018privacy, ahmed2017digital}. With this in mind, as XAI becomes a popular choice for technology developers and user interface designers to integrate within the systems they develop, they should make more effort to be aware of the cultural nuances that could affect the use of these methods. By actively working to understand these nuances, technology will become accessible, effective, and socially responsible for all users, regardless of their respective backgrounds. 

When explaining technical concepts, there exists a variety of barriers that can prevent users from interpreting such information. Technical concepts may also be abstract, containing convoluted theories or ideas that may be challenging for individuals without sufficient background knowledge. Such concepts may also involve technical jargon unfamiliar to users, exacerbating this gap. While the goal of AI explainability has generally been to improve transparency regarding the function of and predictions from AI systems, these benefits are often limited to those with technical backgrounds. The effectiveness of AI explanations is often dependent on factors, including the complexity of the system, the quality of the explanations, and the technical expertise of end users. There are also cases where explanations may be of little or no benefit to users, particularly in situations where interpreting an explanation takes more mental effort than receiving a decision or prediction from an AI system. Miller \cite{miller2019explanation} finds that explanations help improve user understanding of AI but can be counterproductive if they are too complex for users to understand. Explanations are not always complete and can be misleading, limiting their effectiveness for end users. Mittelstadt et al. find that in cases where explanations are technically correct, their respective lack of interpretability by end users can mislead users in understanding functions about AI behavior \cite{mittelstadt2019explaining}. Furthermore, if users do not trust AI systems or the predictions produced by them, the utility of XAI is also reduced. Some  work \cite{cheng2019explaining, lim2009and, zhang2020effect, shin2021effects} finds that while explanations help improve user understanding of AI, the presence of explanations does not always result in enhanced trust in AI systems. Trust is often influenced by users' perceptions of the accuracy and consistency of AI systems and their prior experience interacting with AI systems.

Currently, there exist many barriers to effective XAI. While much technical progress has been made in developing XAI solutions, these systems often do not incorporate findings of human-centered XAI research that advocate for easier interpretability, lower complexity, and  improved transparency. Additionally, many of these systems do not account for technology literacy and cross-cultural differences, especially when it comes to nuances that occur when explaining concepts to end users from locales outside of the West and those with varying levels of technical knowledge. However, prioritizing participatory design and introducing concepts such as intercultural ethics can help address some of these concerns. In the next section, we'll introduce the concept of using an intercultural ethics approach to develop better XAI methods while leveraging participatory methods commonly practiced in fields such as human-computer interaction (HCI) and information and communication technologies for development (ICTD).
\section{Using Intercultural Ethics to Inform XAI}
\label{informingXAI}

To initiate an intercultural ethics approach to XAI, researchers should consider how varying cultural perspectives and values shape the design, development, and implementation of AI systems. There should also be considerations focusing on such perspectives and values that can inform the transparency and accountability of these systems to serve the needs of a diverse set of stakeholders. When we consider the development of XAI from an intercultural ethics approach, researchers should recognize that different cultures have varying expectations for factors such as accountability, transparency, and privacy and that other factors such as trust, power dynamics, and historical experiences shape these expectations. As researchers understand users' expectations, they should also consider how end users in particular contexts value certain aspects of AI systems, such as accuracy and efficiency, and what oversight and control users expect to have over AI decision-making. With this in mind, we outline steps for researchers interested in incorporating intercultural ethics into the development of XAI.

\textbf{Engaging with diverse stakeholders through participatory design}
First, participatory methods should be leveraged by engaging with diverse stakeholders. By engaging end users from different cultural, religious, educational, and professional backgrounds, researchers can expect to gain a broad understanding of the expectations and concerns these users have regarding XAI. Researchers will also gain an understanding of user needs, perspectives, and values. These engagement methods can include co-creation workshops, focus groups, interviews, and user feedback sessions. All of these methods are useful when aiming to tailor XAI to the needs and contexts of specific users. Co-creation workshops bring together stakeholders such as end users, designers, researchers, and developers to generate ideas and design concepts collaboratively. In these workshops, stakeholders can also participate in interactive activities to provide deeper feedback on prototypes developed in these sessions. As these prototypes evolve, hosting focus groups and conducting interviews \cite{adams2008qualititative, kuter2001survey, fredericks2016middle} can be helpful for XAI practitioners to get insights on early iterations of the intended product. User feedback sessions involving members of local communities, especially those from historically marginalized backgrounds, can be further used to solicit feedback on the design and functionality of more advanced prototypes to help progress toward a final product. These sessions can also leverage methods such as interviews, surveys, and usability testing to help practitioners identify areas of improvement. Researchers can create XAI systems that account for user needs and incorporate diverse cultural perspectives by incorporating these participatory methods within their development practices.

\textbf{Studying cultural values and perspectives}
As researchers engage in these participatory design methods, there will be a need to separately study how cultural concepts such as accountability, fairness, privacy, and trust are encoded in the respective context. Early work in this area by \citet{sambasivan2021re} has examined algorithmic fairness in the Indian context, but there has not been much work focused on other regions of the Global South. By extensively studying these factors, researchers can better understand how these cultural nuances can impact the design and development of XAI. Additionally, incorporating local knowledge can provide additional context to XAI systems, making them much more relevant and valuable in places where they are used. There is much work needed to democratize access to AI, and efforts that focus on translating technical jargon into local languages, incorporating cultural nuances, customs, and practices into AI development, increasing training opportunities for AI practitioners, and developing the capacity to support local data collection, data sharing, and cloud storage will be essential to broadening inclusion within AI, eventually impacting how underrepresented users and regions are represented within XAI \cite{okolo2023examining}. Such work cannot be conducted in silos, and AI practitioners will have to intentionally form interdisciplinary collaborations with researchers based in local contexts, researchers with experiences studying cultural nuances, and researchers with domain expertise in the unique interplay between culture and technology. These practices can help avoid ``cultural incongruencies", as coined by Prabhakaran et al. \cite{prabhakaran2022cultural}, where the development and usage of AI systems cause cultural barriers, impose hegemonic classifications, cause safety gaps, violate cultural values, and lead to cultural erasure.

\textbf{Addressing power dynamics to increase transparency and accountability}
As researchers design and develop XAI systems, they should be aware of the power dynamics between them and local communities and work meaningfully to address them. Additionally, researchers must be mindful of power dynamics between certain groups of people in local communities, understanding how social factors such as age, caste, gender, and religion impact communities. To help address power dynamics between themselves and local communities, researchers should implement mechanisms for stakeholder feedback and active input into the design, development, and evaluation processes. Additionally, as explainability becomes incorporated into AI systems, researchers should consider how existing power dynamics may impact the interpretation and use of XAI by local communities. This section has previously highlighted various participatory design methods that practitioners can leverage, which can help address issues of power dynamics. Using these methods, researchers can work towards inclusive XAI systems that prioritize transparency and accountability in their decision-making processes. To ensure these systems are responsive to stakeholder concerns and expectations, practitioners can also implement mechanisms to audit and monitor XAI systems. XAI can augment standard auditing practices in AI \cite{raji2020closing, cobbe2021reviewable} by ensuring that the outcomes of audits are interpretable to stakeholders across a spectrum of technical knowledge. AI has been shown to exacerbate disability-based discrimination \cite{morris2020ai}, and XAI can further amplify such discrimination if researchers fail to incorporate inclusive measures within AI development. By addressing power dynamics, prioritizing transparency and accountability, and leveraging accessibility measures in HCI and AI \cite{stephanidis1998universal, obrenovic2007universal, wickramasinghe2020trustworthy}, practitioners can make XAI more accessible to populations within the Global South who have limited experience with digital technologies and low exposure to AI.   

\textbf{Conducting human-centered evaluations}
Prior work within XAI has advocated for robust evaluations of XAI in low-resource contexts, especially those situated within the Global South \cite{okolo2022making}. Low-resource contexts, regardless of location, provide a variety of challenges that can hamper the effective integration and use of XAI in these domains. Extenuating factors include a lack of data sources, limited internet connectivity, low AI/digital literacy, smaller concentrations of AI practitioners, and a shortage of computing resources needed to train and evaluate these systems for real-world use. In the Global South, these factors can be compounded by cultural nuances and societal structures that set different expectations for who engages with, uses, and governs technology. To proactively understand and address these challenges, researchers should rigorously test XAI in real-world settings to ensure they are effective, fair, and reliable. Our advocacy for leveraging participatory methods within the design process also translates to the evaluation process. To create effective environments for evaluating XAI, practitioners should also intentionally partner with local organizations and institutions to conduct user testing and evaluations. Collaborations with such partners leverages their respective domain expertise and experience interacting with local communities and can also help quickly identify issues for researchers to address.
\section{Discussion}
\label{discuss}

Comprehensively, an intercultural ethics approach to XAI promotes diverse cultural perspectives and values while actively working to incorporate these considerations into the design, development, and deployment of AI systems. As research within the field of human-centered XAI continues to expand, practitioners can take advantage of the recommendations provided within this paper to develop effective and equitable systems that meet the needs and expectations of users. However, these recommendations are not to be taken without examining the risks that XAI and AI, in general, pose to communities. Although AI has the potential to provide many benefits to communities, the harms of these technologies have rightly overshadowed this potential. Introducing XAI into flawed AI systems can exacerbate these ills, resulting in a detrimental user experience that harms user agency, violates cultural and personal values, and decreases trust in AI systems. In cases where (X)AI systems are used for high-stakes decisions in domains like healthcare or lawmaking, negative impacts made by AI-assisted decision-making tools could be irrevocable, permanently harming people subjected to these systems. With this in mind, AI practitioners should critically reflect on their role as researchers, designers, and developers to understand their motivations for developing XAI and to what ends it serves their interests over those of populations who will be affected by the systems as they reach the real world.

\textbf{Limitations} 
We posit an intercultural ethics approach to XAI as only part of a solution for AI practitioners to improve XAI development. However, we acknowledge the existing challenges and barriers in AI research and the implications of using (X)AI to address societal problems. AI should not be seen as the solution to all issues, and feasible approaches that don't leverage AI should also be considered as practitioners outline approaches to their research problems. While this paper primarily focuses on XAI, intercultural ethics can be applied to other domains of artificial intelligence, machine learning, and even technology as a whole. We encourage researchers interested in building upon this work, which we characterize as \textit{intercultural XAI}, to apply our guidelines in various contexts and experiment with augmentations to improve the robustness of this approach and its applicability to domains outside of AI.  
\section{The Future of Intercultural XAI}
\label{future}

An intercultural ethics approach to XAI is only one approach to improving the cultural competency of XAI methods. As researchers continue building and evaluating XAI methods, this research may inspire other avenues of exploration. Identifying complementary approaches to intercultural ethics is a promising area of further research, as Prabhakaran et al. state: "culture is not a static variable that can be easily encoded into a technology, rather a complex and dynamic system that is constantly being transmitted and transformed" \cite{prabhakaran2022cultural}. We encourage parallel and tangential approaches to our work and look forward to this subdomain growing even further. We end this article outlining a set of questions for researchers to contemplate as they assess the prospect of integrating intercultural ethics:

\begin{itemize}
  \item \textbf{How can we leverage intercultural XAI to examine best practices for incorporating cultural values and perspectives into XAI development?} To understand what intercultural XAI looks like in practice, interdisciplinary collaborations that leverage the work presented in this paper could produce novel findings, providing rich contributions to various fields, including computing, digital humanities, sociology, and anthropology. 
  \item \textbf{What lessons can (X)AI practitioners learn from practices of colonial exploitation on modern-day AI to improve their research and development practices?} To complement the literature on decolonial AI \cite{mohamed2020decolonial, mhlambi2023decolonizing, birhane2020algorithmic}, researchers can examine these perspectives in tandem with intercultural XAI to understand better practices to foster inclusion in the development pipelines of AI systems and develop more user-interpretable methods.
  \item \textbf{What impact does intercultural XAI have on the efficacy and adoption of XAI?} As XAI continues to be developed and reconstructed, evaluation is essential to this process. To understand the value of intercultural XAI, researchers may consider developing novel evaluation metrics that provide sufficient context to model performance and user experience.
\end{itemize}
\section{Conclusion}
\label{conclusion}

This paper introduces the concept of examining AI explainability through the lens of intercultural ethics. We highlight current work in XAI, noting the significant progress made towards improving interpretation methods, discuss current barriers to explaining AI, and outline ways to leverage intercultural ethics to inform the design, development, and implementation of XAI. We acknowledge that XAI is not the panacea to completely address the entirety of issues plaguing AI and understand that there are use cases where explanations are less helpful or not needed for a respective context. By considering and implementing an intercultural ethics approach to XAI development, we hope that researchers can improve the efficiency and inclusivity of XAI while ensuring that their tools serve the needs of a diverse user base. As AI continues to evolve as a field and transcend borders, novel approaches such as these will be needed to help mitigate negative impacts and broaden the potential of AI as a force for good.


\bibliographystyle{ACM-Reference-Format}
\bibliography{biblio}

\appendix

\end{document}